\documentclass[manuscript]{acmart}
\usepackage{fancyvrb}
\usepackage{tabularx}

\AtBeginDocument{%
  }

\setcopyright{acmlicensed}
\copyrightyear{2024}
\acmYear{2024}
\acmDOI{XXXXXXX.XXXXXXX}

\begin{document}

\title{Towards Pedagogical LLMs with Supervised Fine Tuning for Computing Education}

\author{Alexandra Vassar}
\orcid{0000-0001-8856-2566}
\author{Jake Renzella}
\orcid{0000-0002-9587-1196}
\author{Emily Ross}
\orcid{0009-0007-5566-7493}
\author{Andrew Taylor}
\orcid{0000-0003-4741-0069}
\affiliation{%
  \institution{The University of New South Wales}
  \city{Sydney}
  \country{Australia}
}

\renewcommand{\shortauthors}{Vassar et al.}

\begin{abstract}
This paper investigates supervised fine-tuning of large language models (LLMs) to improve their pedagogical alignment in computing education, addressing concerns that LLMs may hinder learning outcomes. The project utilised a proprietary dataset of 2,500 high quality question/answer pairs from programming course forums, and explores two research questions: the suitability of university course forums in contributing to fine-tuning datasets, and how supervised fine-tuning can improve LLMs' alignment with educational principles such as constructivism. Initial findings suggest benefits in pedagogical alignment of LLMs, with deeper evaluations required.

\end{abstract}



\received{20 February 2007}
\received[revised]{12 March 2009}
\received[accepted]{5 June 2009}

\maketitle

\section{Introduction}
Developments in generative AI, led by the widespread commercial release of OpenAI's large language models (LLMs) sparked a flurry of applications in computing education, promising to improve learning outcomes and reduce attrition, with a focus on AI-generated programming error message explanations (PEMs) \cite{Taylor2024DccModels, Prather2023TheEducation, Liu2024TeachingEducation, Leinonen2023UsingMessages, Kimmel2024EnhancingAI, Wang2024ACS1, Liffiton2023CodeHelp:Classes}. Leinonen et al. \cite{Leinonen2023UsingMessages} found that Codex produced novice-understandable PEMs for common Python errors. Taylor et al. \cite{Taylor2024DccModels} found 83\% of AI-generated PEMs to be accurate. Students also self-report LLM error explanations to be helpful \cite{Liu2024TeachingEducation}.

Despite reported benefits, commercially available LLMs are aligned to be helpful assistants, which may oppose key tenets of education. Constructivism, one dominant pedagogical theory, states that learners build knowledge by doing rather than being told \cite{Ben-Ari2001}, and is supported by cognitive psychology literature, which states acquiring knowledge is a function of time and conscious effort \cite{Sweller2023CognitiveLearn}. Commercially available LLMs contradict these tenets, displaying a propensity to provide students with solutions despite being instructed otherwise \cite{Taylor2024DccModels, Prather2024TheProgrammers}; potentially harming learning by reducing self-efficacy and grades \cite{Padiyath2024InsightsCourse, Dalalah2023TheChatGPT, Denny2024ComputingAI}.

This paper presents our process and initial findings of fine-tuning ChatGPT3.5 to improve pedagogical alignment within computing education. The design of the fine-tuned model, now deployed to over 600 programming students in a large Australian university, was guided by the following research questions:

\textbf{RQ1}: How effectively can university course forums contribute to fine-tuning datasets?

\textbf{RQ2}: How can supervised fine-tuning better align large language models with pedagogical behaviours?

\section{Debugging C Compiler and AI Extension}
The Debugging C Compiler (DCC) is an educationally-focused C compiler built upon the Clang implementation \cite{Taylor2023FoundationsCompiler}. DCC automates industry-grade dynamic and static memory analysis tools such as Valgrind and AddressSanitizer, simplifying complex tools. An extension to DCC, \textit{DCC Help} embeds an AI-generated error explainer system into the tool. Source code, error messages, and GDB memory stack frame information are utilised to produce better error explanations tailored to novices \cite{Taylor2024DccModels}. Built upon ChatGPT 3.5 Turbo and used over 450,000 times, it is distinct from other tools as it is abstracted and integrated into DCC itself, meaning novice students do not need to play any role in prompt construction, or passing in source code and error messages.


\begin{table*}[h]
    \centering
    \tiny
    \caption{Comparison of error explanations provided by GPT-3.5 and our fine-tuned model (FT2)}
    \begin{tabularx}{\textwidth}{|>{\raggedright\arraybackslash}X|>{\raggedright\arraybackslash}X|}
        \hline
        \textbf{Explanation provided by GPT-3.5} & \textbf{Explanation provided by our fine-tuned model (FT2)} \\
        \hline
        \multicolumn{1}{|>{\arraybackslash}p{0.55\textwidth}|}{\texttt{The error message indicates that some variables (`row`, `col`, `row\_end`, `col\_end`, `row\_start`, `col\_start`, `letter`) are being used without being initialized with a value before being used. This can cause undefined behavior in your program. To fix this error, make sure to initialize these variables with some default or initial values before using them in your program. This will ensure that the variables have valid values when they are being used in different parts of your program. Remember to always initialize variables before using them to prevent issues like this.}} &
        \multicolumn{1}{>{\arraybackslash}p{0.4\textwidth}|}{\texttt{Heya! The warning is exactly right; you're using these variables without initializing them. Anything the user types in the `while` loop will use these variables, but if you're just setting them after, they won't be initialized outright. How might you make sure that every variable is initialized no matter what?}} \\
        \hline
    \end{tabularx}
    \label{tab:error-explanations}
\end{table*}

\section{Methodology, Initial Results and Discussion}
A proprietary dataset of high-quality question/answer pairs across five terms of our programming course forums were sourced from verified tutors during paid hours, and confirmed to be verified responses. This dataset was then used to fine-tune OpenAI's ChatGPT 3.5 model (then state-of-the-art), and made available to students via DCC Help.

The following, iterative process was conducted to produce the final fine-tune (FT2). Steps were repeated when necessary, gradually refining the model until it met acceptance criteria.

\begin{enumerate}
    \item[Step 1] \textbf{Data Collection}: Simple API to extract raw Q/A pairs from our programming course forums.
    \item[Step 2] \textbf{Data Cleansing}: Automated filtering (URLs, proper nouns), grammatical corrections using ChatGPT 4o.
    \item[Step 3] \textbf{Manual filtering and quality control}: Filtering for the inclusion criteria (outlined below).
\end{enumerate}

Data cleansing was critical, as many responses were informal, containing grammatical issues which impacted fine-tuning. ChatGPT-4o was used to correct grammatical issues such as incorrect spelling, typos, adding capitalisation and adding code-block fences.

We evaluated an early fine-tuned model (FT1), but it suffered from prohibitive quality issues. Manual filtering in Step 3 was therefore motivated: five tutors each reviewed 500 of 2500 randomly assigned Q/A pairs, applying the following inclusion criteria: \textbf{a)} answers must be correct, helpful, and self-contained; \textbf{b)} provide suggestions rather than solutions; \textbf{c)} have formal tone without being dismissive; \textbf{d)} include code blocks as examples only; \textbf{e)} avoid names, specific assignments, or lab exercises; \textbf{f)} focus on programming language understanding, bugs, and style.

There were 528 pairs that fit all the criteria described above, constituting 21\% of the dataset. While time-consuming, this process was vital in improving the dataset and resulted in a far improved fine-tune (FT2). Fifty real student questions were then evaluated with GPT-3.5, GPT-4o, and FT2. The FT2 results featured a more informal language style, in line with that of our tutors. Compared to the instructive tone of GPT-3.5, where solutions are plainly stated and sometimes given, FT2 Socratically prompts the student to consider a particular approach (\autoref{tab:error-explanations}). The FT2 responses are concise, while conveying similar information. Comparatively, GPT-4o responses are verbose: overwhelming the terminal environment. Future work will rigorously measure FT2 quality utilising the methodology presented in \cite{Taylor2024DccModels}.

\bibliographystyle{ACM-Reference-Format}
\bibliography{main_matter}

\end{document}